\documentclass{article}
\usepackage{ijcai26}

\usepackage{times}
\usepackage{soul}
\usepackage{url}
\usepackage{hyperref}
\hypersetup{hidelinks}
\usepackage[utf8]{inputenc}
\usepackage[small]{caption}
\usepackage{graphicx}
\usepackage{amsmath}
\usepackage{amssymb}
\usepackage{amsthm}
\usepackage{booktabs}
\usepackage{algorithm}
\usepackage{algorithmic}
\usepackage{multirow}
\usepackage{tikz}
\usetikzlibrary{positioning, shapes.geometric, arrows.meta, fit, calc}

\title{Addressing Market Regime Changes and Heavy-Tailed Returns in Portfolio Optimization via Bayesian VAR and Elliptical Black-Litterman}

\author{
    Daniil Mikriukov$^{1,2}$\and
    Ruoyu Sun$^{2}$\and
    Angelos Stefanidis$^{2}$\and
    Jionglong Su$^{2}$\and
    Zhengyong Jiang$^{2}$
    \affiliations
    $^1$Department of Computer Science, University of Liverpool, Liverpool, United Kingdom\\
    $^2$School of AI and Advanced Computing, Xi'an Jiaotong-Liverpool University, Suzhou, China
    \emails
    D.Mikriukov@liverpool.ac.uk,
    fiez0303@outlook.com,
    angelos.stefanidis@xjtlu.edu.cn,
    jionglong.su@xjtlu.edu.cn,
    zhengyong.jiang02@xjtlu.edu.cn
}

\begin{document}

\maketitle

\begin{abstract}
Deep reinforcement learning (DRL) frameworks for portfolio optimization have shown promise for their ability to learn allocation rules dynamically from market data. However, these models fail to account for fat-tailed returns, which characterize actual market behavior with more frequent extreme events. Furthermore, historical data is treated homogeneously, without accounting for temporal importance, leading models to fail during regime changes. We propose a new BAVAR-BLED algorithm that combines methods derived from Bayesian-Averaging Vector Autoregressive (BAVAR) and the Black-Litterman model using Elliptical Distributions (BLED) within a TD3 architecture. BAVAR captures a set of vector autoregressive representations that consider multi-scale temporal features, enabling adaptive allocation decisions based on regime-aware estimates of return expectations and dispersion matrices. These estimates serve as prior inputs to BLED, a model that uses Student's $t$-distributions, allowing for more realistic fat tail return estimates. The BAVAR-BLED algorithm uses transformer networks for view construction and CNNs for risk-aversion estimates, which modify dynamic allocation decisions based on market conditions. An evaluation of 29 Dow Jones Industrial Average constituents over a decade-long market period shows that BAVAR-BLED significantly outperforms state-of-the-art methods, achieving Sharpe and Sortino ratios of 1.72 and 2.70, respectively, and total returns of 57.26\%.
\end{abstract}

\section{Introduction}
\label{sec:introduction}

Portfolio optimization (PO) is a well-known problem in quantitative finance, where investors have to balance expected returns against associated risks. One of the earliest works in this field belongs to \citeauthor{markowitz1952} with his mean-variance framework \shortcite{markowitz1952}. While it remains quite influential in PO, its high sensitivity to estimation errors in the sample mean and covariance matrix often leads to unstable portfolio allocations, particularly during market stress when correlations increase and historical estimates become unreliable.

Recent advances in deep reinforcement learning (DRL) demonstrate great potential for adaptive portfolio management, as agents can directly learn to make allocation decisions purely from market data by continuously capturing and adapting to market conditions \cite{jiang2017deep,fischer2018deep}. However, the majority of existing DRL frameworks for PO fail to account for the well-documented fat-tailed nature of financial returns \cite{ellip3,ellip2,ellip1}, which leads to an underestimation of tail risks and potentially significant losses during market turbulence.


The Black-Litterman model addresses the mean-variance's estimation error by combining market equilibrium returns with investor views inside of the Bayesian framework \cite{black1992global}. More recent work by \citeauthor{xiao2015black} further generalized this framework to elliptical distributions to more realistically model heavy-tailed returns \shortcite{xiao2015black}. In our prior work \cite{mikriukov2025bled}, we introduced the Black-Litterman under Elliptical Distributions (BLED) framework, which embeds this elliptical extension into a transformer-based DRL agent and demonstrated improved risk-adjusted performance compared to both traditional methods and standard DRL models. However, BLED still faces a fundamental limitation during regime changes when computing the market prior (the expected return vector derived from market equilibrium) and dispersion matrix (the generalization of covariance that captures return co-movements under elliptical distributions) by treating the historical data uniformly, whereas recent observations should carry more weight than distant ones.

To address this limitation, we propose a novel BAVAR-BLED framework that integrates Bayesian-Averaging Vector Autoregressive (BAVAR) methods with our BLED framework inside a DRL agent, where the first component is based on the methodology of \citeauthor{anderson2022bavar} and maintains an ensemble of heterogeneous VAR models that are continuously updated via Bayesian learning \shortcite{anderson2022bavar}. This ensemble replaces the static historical estimates used in standard BLED with adaptive ones for expected returns and dispersion matrix that respond to regime changes. Furthermore, to capture multi-scale temporal dynamics and to incorporate information from daily, weekly and monthly return patterns, our framework utilizes Heterogeneous Autoregressive (HAR) features \cite{corsi2009har}.

The four key contributions of this paper are as follows:
\begin{itemize}
    \item To the best of our knowledge, we are first to propose the integration of BAVAR methodology with our BLED framework inside a DRL agent, providing adaptive market prior and dispersion estimation that responds to regime changes.
    \item We also show the significant improvement of the portfolio performance during volatile market periods by the addition of adaptive BAVAR-derived priors in comparison with static ones.
    \item Validation on 29 stocks from the DJIA from 2014 to 2024 is conducted, where we demonstrate the outperformance of state-of-the-art methods by the proposed BAVAR-BLED framework, which achieves superior risk-adjusted performance (Sharpe ratio of 1.72 and Sortino ratio of 2.70).
    \item Lastly, we conduct comprehensive ablation studies, where the individual contribution of each component to overall performance is demonstrated.
\end{itemize}

Our framework operates under standard assumptions for portfolio optimization: trading at adjusted closing prices, fractional shares permitted, zero market impact, immediate execution without slippage, and a transaction cost of 0.25\% per rebalance, which is calibrated for a \$100,000 USD portfolio trading liquid US equities \cite{frazzini2018trading}. Short selling is also permitted in our simulated environment.

\section{Related Work}

\subsection{Deep Reinforcement Learning for Portfolio Optimization}

Deep reinforcement learning has gained significant attention for portfolio management due to its ability to learn complex decision policies from data \cite{gao2020application,song2023deterministic}. \citeauthor{jiang2017deep} introduced the Ensemble of Identical Independent Evaluators (EIIE) topology, using convolutional neural networks to process price tensors and generate portfolio weights \shortcite{jiang2017deep}. Subsequent work has explored various architectures including LSTM networks \cite{fischer2018deep}, attention mechanisms \cite{yang2023deep}, and multi-agent systems \cite{shavandi2022multi,huang2022mspm}. The Twin Delayed Deep Deterministic Policy Gradient (TD3) algorithm \cite{fujimoto2018addressing} has also become quite a popular choice for continuous action spaces in portfolio optimization due to its stability and reduced overestimation bias. Besides that, recent applications include risk-controlled portfolios \cite{trc2,choudhary2025risk} and hierarchical approaches \cite{gao2021framework}. Transformer architectures have also shown promise for capturing long-range dependencies in financial time series \cite{yang2023deep,liu2024revolutionising,zhang2024finbpm}. Despite these advances, most DRL methods fail to account for the heavy-tailed nature of financial returns, leading to underestimation of tail risks.

\subsection{Black-Litterman Model and Extensions}

To address the fat-tailed nature of financial returns, \citeauthor{xiao2015black} extended the Black-Litterman model \cite{black1992global} to elliptical distributions, thus allowing for more realistic modeling of heavy-tailed returns through distributions such as the Student's $t$ \shortcite{xiao2015black}.

Some recent works have integrated the Black-Litterman framework with deep learning. For instance, \citeauthor{sun2024combining} combined transformer-based DRL with the Black-Litterman model for portfolio optimization \shortcite{sun2024combining}. Our prior work \cite{mikriukov2025bled} further extended this direction by incorporating elliptical distributions (BLED) and combining transformer view generation with CNN-based risk aversion estimation. \citeauthor{zhu2024enhancingportfoliooptimizationtransformergan} had slightly different vision and explored GAN-based view generation within the Black-Litterman framework \shortcite{zhu2024enhancingportfoliooptimizationtransformergan}. However, these approaches treat historical data uniformly when computing market priors, which becomes rather problematic during regime changes when recent observations should carry more weight.

\subsection{Bayesian VAR Methods in Finance}

Vector autoregressive models have a long history in financial econometrics for modeling multivariate time series. Bayesian approaches may be used to further improve them by providing natural uncertainty quantification and incorporating prior knowledge about coefficient stability. Using this fact, \citeauthor{anderson2022bavar} developed the BAVAR methodology, which maintains an ensemble of VAR models with different hyperparameters and uses Bayesian model averaging to produce forecasts, thus making it well-suited for non-stationary environments where model parameters may shift over time \shortcite{anderson2022bavar}. This approach further benefits from the Heterogeneous Autoregressive (HAR) model of \citeauthor{corsi2009har} that captures multi-scale temporal dynamics by including lagged values at different frequencies \shortcite{corsi2009har}. HAR features were originally developed for realized volatility forecasting, but have demonstrated their effectiveness for modeling various financial time series. Furthermore, the combination of these features with Bayesian VAR methods enables models to capture both short-term fluctuations and longer-term trends. Unlike regime-switching models that require a predefined number of regimes, BAVAR handles regime changes implicitly through its ensemble structure, outputting adaptive priors that respond to structural shifts.

\subsection{Research Gap}

Existing work in DRL-based portfolio optimization faces two fundamental limitations. First, most approaches fail to account for the well-documented fat-tailed nature of financial returns \cite{ellip3,ellip2,ellip1}. While our BLED framework \cite{mikriukov2025bled} addresses this by using elliptical distributions, it inherits a second limitation: reliance on static historical estimates for market priors, which becomes quite troublesome during regime changes when asset relationships evolve.

Our BAVAR-BLED framework addresses both of the aforementioned gaps within a unified architecture, where the key innovation lies in the integration of Bayesian VAR methods with the elliptical Black-Litterman model, which provides adaptive, regime-aware prior estimates that respond to changing market conditions while modeling heavy-tailed returns more realistically.

\section{Methodology}

\subsection{Framework Overview}

\begin{figure*}[t]
\centering
\includegraphics[width=\textwidth]{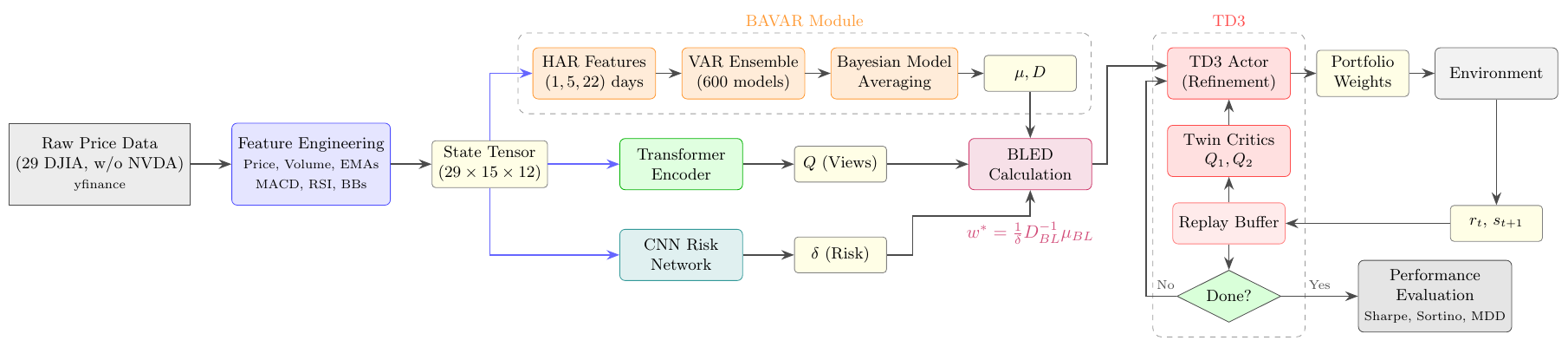}
\caption{BAVAR-BLED framework architecture. Historical price data is fed into feature engineering to produce state tensors. Three parallel branches process the state: (1) the BAVAR module extracts HAR features and maintains an ensemble of VAR models to produce adaptive prior estimates $\mu$ and dispersion $D$; (2) the Transformer encoder generates investor views $Q$; (3) the CNN estimates risk aversion $\delta$. These outputs feed into the BLED calculation, which computes Black-Litterman optimal weights under elliptical distributions. The TD3 actor refines these weights, and twin critics estimate Q-values for policy optimization.}
\label{fig:architecture}
\end{figure*}

Figure~\ref{fig:architecture} illustrates the overall system design of the BAVAR-BLED framework, which includes data preprocessing, three parallel processing branches, Black-Litterman optimization, and TD3 reinforcement learning. Here the raw price data for 29 DJIA constituent stocks is fetched from yfinance and processed through feature engineering to extract 12 features per asset: adjusted close price, volume, five exponential moving averages (EMA$_{10,20,50,100,200}$), MACD with signal line, RSI, and Bollinger Bands, which produces a state tensor of shape $(29 \times 15 \times 12)$ that represent 29 assets over a 15-day window with 12 features each. The state tensor is then fed into three parallel branches: (1) the BAVAR module extracts HAR features and maintains an ensemble of 600 VAR models to produce adaptive prior estimates $\mu$ and dispersion matrix $D$; (2) a transformer encoder generates investor views $Q$; and (3) a CNN estimates the risk aversion parameter $\delta$. These outputs are then used in the BLED calculation, which computes Black-Litterman optimal weights under elliptical distributions. Lastly, the model undergoes the TD3 training loop through interactions with the environment.

\subsection{MDP Formulation} \label{sec:mdp}

We formulate the PO problem as a Markov Decision Process (MDP) with continuous state and action spaces.

\paragraph{State Space.}
At each time step $t$, the state $s_t \in \mathbb{R}^{n \times w \times f}$ represents a tensor of historical observations, where $n=29$ is the number of assets, $w=15$ is the lookback window in trading days, and $f=12$ is the number of features per asset. Features include adjusted close price, trading volume, and exponential moving averages (EMA$_{10,20,50,100,200}$), MACD with signal line, RSI, and Bollinger Bands.

\paragraph{Action Space.}
The action $a_t \in \mathbb{R}^{n+1}$ represents target portfolio weights for $n$ assets plus a cash position. Also to enable both long and short positions, we employ absolute-value normalization to weights: $\sum_{i=1}^{n+1} |w_i| = 1$ with $w_i \in [-1, 1]$, where negative weights represent short positions. 

\paragraph{Reward Function.}
The reward is the logarithmic portfolio return after transaction costs:
\begin{equation}
    r_t = \log(V_t + \varepsilon) - \log(V_{t-1} + \varepsilon),
\end{equation}
where $V_t = V_{t-1} \cdot (1 + r_p) \cdot (1 - \mu_t)$ is the portfolio value after returns and costs, $r_p = \sum_i w_i \cdot r_{i,t}$ is the weighted portfolio return, $\mu_t = c \cdot \sum_i |w_{i,t} - w_{i,t-1}|$ is the transaction cost proportional to turnover with $c = 0.0025$, and $\varepsilon = 10^{-8}$ ensures numerical stability.

\subsection{BAVAR Component}

BAVAR maintains an ensemble of VAR models that update continuously with new data, thereby providing adaptive estimates of expected returns and dispersion matrix.

\paragraph{HAR Feature Extraction.}
Following \citeauthor{corsi2009har} \shortcite{corsi2009har}, we construct multi-scale temporal features from asset returns:
\begin{equation}
    x_{t-1} = \left[1, \bar{r}^{(d)}_{t-1}, \bar{r}^{(w)}_{t-1}, \bar{r}^{(m)}_{t-1}\right]^\top,
\end{equation}
where superscripts $d$, $w$, and $m$ denote daily, weekly, and monthly scales respectively. Specifically, $\bar{r}^{(d)}_{t-1}$ is the previous day's return, while $\bar{r}^{(w)}_{t-1}$ and $\bar{r}^{(m)}_{t-1}$ represent returns averaged over the previous 5 and 22 trading days. These features capture both short-term momentum and medium-term mean reversion patterns.

\paragraph{VAR Model Specification.}
We assume that asset returns follow a time-varying coefficient VAR:
\begin{equation}
    r_t = B_t x_{t-1} + \epsilon_t,
\end{equation}
where $r_t \in \mathbb{R}^n$ are excess returns for $n$ assets, $B_t$ is the true (unobserved) coefficient matrix that may shift when regimes change, and $\epsilon_t \sim \mathcal{N}(0, \Sigma_t)$. Each model $m$ in the ensemble maintains a posterior estimate $\bar{B}_{m,t}$ of $B_t$ based on its assumed regime structure and hyperparameters.

\paragraph{Bayesian Update Equations.}
Following \citeauthor{anderson2022bavar} \shortcite{anderson2022bavar}, we update each model's sufficient statistics using the following recursions:
\begin{align}
    \Lambda_{m,t} &= \Lambda_{m,t-1} + \frac{e_{m,t} e_{m,t}^\top}{1 + x_{t-1}^\top \Phi_{m,t-1} x_{t-1}} \label{eq:lambda} \\
    \nu_{m,t} &= \nu_{m,t-1} + 1 \label{eq:nu} \\
    \Phi_{m,t} &= \left(\Phi_{m,t-1}^{-1} + x_{t-1} x_{t-1}^\top\right)^{-1} \label{eq:phi} \\
    \bar{B}_{m,t} &= \Phi_{m,t} \left[\Phi_{m,t-1}^{-1} \bar{B}_{m,t-1} + x_{t-1} r_t^\top\right], \label{eq:bbar}
\end{align}
where $e_{m,t} = r_t - \bar{B}_{m,t-1} x_{t-1}$ is the prediction error, $\Lambda_{m,t}$ accumulates the sum of squared errors, $\nu_{m,t}$ tracks degrees of freedom, $\Phi_{m,t}$ is the precision matrix for coefficients, and $\bar{B}_{m,t}$ is the posterior mean estimate of the true coefficient matrix $B_t$. Each model $m$ is initialized with diffuse priors: $\nu_{m,0} = n + 2$, $\Lambda_{m,0} = \bar{\sigma}^2 I_n$ where $\bar{\sigma}^2$ is the sample variance, and $\bar{B}_{m,0}$ sets the constant term to the sample mean with remaining coefficients at zero. The precision matrix $\Phi_{m,0} = \text{diag}(\alpha_i, \beta_i I_{k-1})$ varies across models with $\alpha_i \in \{0.01, 0.1, 1\}$ and $\beta_i \in \{1, 10, 100\}$, creating 9 hyperparameter combinations that ensure ensemble diversity. Updates proceed for $t = 1, 2, \ldots, T$.

\paragraph{Model Weight Update.}
The weight of each model is updated via Bayes' rule based on predictive likelihood:
\begin{equation}
    P(m|F_t) = \frac{P(r_t|m, F_{t-1}) P(m|F_{t-1})}{\sum_{m' \in \mathcal{M}_t} P(r_t|m', F_{t-1}) P(m'|F_{t-1})},
\end{equation}
where $F_t$ denotes all information up to time $t$, $P(r_t|m, F_{t-1})$ is the predictive density under model $m$, and the summation in the denominator runs over all models $m'$ in the active set $\mathcal{M}_t$ to ensure weights sum to one.

\paragraph{Bayesian Model Averaging.}
The BAVAR predictions combine individual model forecasts weighted by posterior model weights. Each model $m$ produces mean and covariance predictions:
\begin{align}
    \mu_m &= \bar{B}_m^\top x_{t-1} \label{eq:mu_m} \\
    \Sigma_m &= \frac{\Lambda_m}{\nu_m - n - 1}, \label{eq:sigma_m}
\end{align}
which are combined via:
\begin{align}
    \mu &= \sum_m P(m|F_t) \mu_m \label{eq:mu_bavar} \\
    D &= \sum_m P(m|F_t) \left[\Sigma_m + \mu_m \mu_m^\top\right] - \mu \mu^\top, \label{eq:sigma_bavar}
\end{align}
where $\mu$ and $D$ denote the prior mean and dispersion matrix used in the subsequent BLED integration.

%

\paragraph{Ensemble Dynamics.}
The ensemble initializes new models with diffuse priors as specified above during each update call, which then accumulate weight during Bayesian updating while processing market observations. Models with poor predictive power receive lower weights and contribute less to the averaging process, while well-performing models dominate it, allowing the ensemble to implicitly adapt to regime changes without explicit regime detection. 

\subsection{BLED Integration}

The Black-Litterman model under Elliptical Distributions combines the BAVAR-derived prior distribution of expected returns and dispersion with transformer-generated views and CNN-produced risk aversion to produce posterior estimates of expected returns and dispersion.

\paragraph{Elliptical Distribution Framework.}
We model asset returns using the Student's $t$-distribution, a member of the elliptical family with heavier tails than the normal distribution, which has the density generator:
\begin{equation}
    f_X(x; \mu, D) = |D|^{-1/2} g_n\left((x - \mu)^\top D^{-1} (x - \mu)\right),
\end{equation}
where $\mu$ is the location parameter, $D$ is the dispersion matrix, and $g_n$ is the density generator specific to the $t$-distribution with $\nu$ degrees of freedom.

\paragraph{Black-Litterman Posterior.}
The posterior expected return and dispersion matrix are computed as:
\begin{align}
    \mu_{\text{BL}} &= \mu + \left[(\tau D)^{-1} + P^\top \Omega^{-1} P\right]^{-1} P^\top \Omega^{-1} (Q - P\mu), \label{eq:mu_bl} \\
    D_{\text{BL}} &= D - D P^\top (Q + P D P^\top)^{-1} P D, \label{eq:d_bl}
\end{align}
where $\mu$ and $D$ are the prior mean and dispersion matrix from BAVAR (Equations~\eqref{eq:mu_bavar} and \eqref{eq:sigma_bavar}), $P$ is the view matrix (identity matrix for absolute views), $Q$ contains the view magnitudes generated by the transformer, $\Omega$ is the view uncertainty matrix, and $\tau$ is a scaling parameter that controls the confidence in the prior.

\paragraph{Transformer View Generation.}
A transformer component generates investor views on expected returns, while positional encoding is omitted since the 15-day window already captures temporal structure, thus allowing it to focus on asset correlations. Within this module, multi-head self-attention layers and feed-forward networks produce the view vector $Q \in \mathbb{R}^n$ representing expected returns for each asset.

\paragraph{CNN Risk Aversion Estimation.}
A convolutional neural network processes the state tensor to estimate a state-dependent risk aversion parameter $\delta$. This network applies convolutional layers to extract patterns from price and volume data, followed by fully connected layers that output a scalar bounded within a predefined range. This component allows the model to increase risk aversion during volatile periods and decrease it during stable conditions. Finally, the optimal portfolio weights are computed as:
\begin{equation}
    w^* = \frac{1}{\delta} D_{\text{BL}}^{-1} \mu_{\text{BL}}. \label{eq:optimal_weights}
\end{equation}

\subsection{TD3 Training}

Given the MDP formulation with continuous state and action spaces (Section~\ref{sec:mdp}), we employ the TD3 algorithm to learn a policy that maximizes cumulative log returns.

\paragraph{Actor-Critic Architecture.}
The TD3 algorithm uses twin Q-networks ($Q_1$, $Q_2$) to reduce overestimation bias and employs delayed policy updates for stability, while the actor network takes the BLED-derived weights from Equation~\eqref{eq:optimal_weights} as input and learns refinements under the assumptions and constraints specified in Section~\ref{sec:introduction}. Experiences are stored in a replay buffer, from which mini-batches are sampled for training of the critics and the actor.

\paragraph{Training Loop.}
At each environment step, the agent observes the state, computes BLED weights, applies actor refinement, and executes the portfolio allocation. The environment returns the reward and next state, which are stored in the replay buffer. The critics are updated by minimizing the Bellman error, and the actor is updated via deterministic policy gradient with delayed updates. Training continues until convergence, then performance is evaluated using the Sharpe ratio, Sortino ratio, and maximum drawdown.

\paragraph{BAVAR Update Strategy.}
BAVAR models are updated during environment steps, while the TD3 algorithm samples batches from the replay buffer for policy and value function updates. This separation ensures that the BAVAR ensemble processes observed returns in their natural sequential order, preventing overfitting to replayed trajectories and maintaining temporal integrity of the Bayesian updates. Lastly, to enable parallel execution without blocking the DRL loop, the actor network uses cached BAVAR priors that are refreshed asynchronously in a separate computational thread.

\paragraph{Computational Optimization.}
The computational bottleneck is introduced by using synchronous updates for the BAVAR ensemble at each environment step, requiring approximately 2600 seconds per training episode. We developed an adaptive hybrid wrapper that performs BAVAR updates asynchronously at reduced frequency. Empirical testing showed that 12 updates per year, corresponding to approximately monthly updates or every 21 trading days, provides a good balance between computational cost and model accuracy. This addition reduced the episode training time to approximately 750 seconds and allowed for more efficient hyperparameter optimization.

\section{Experiments}

\subsection{Experimental Setup}

\paragraph{Dataset.}
We evaluate BAVAR-BLED across 29 Dow Jones Industrial Average (DJIA) constituent stocks, excluding NVIDIA due to its outlier performance, which would distort the comparative analysis. The data spans January 2014 to December 2024, providing over 10 years of daily observations including multiple market regimes such as the COVID-19 crash and recovery, interest rate cycles, and sector rotations.

We use a 60/20/20 chronological split for training, validation, and testing, resulting in 1640 training days, 547 validation days, and 547 test days. Each trading day, the agent observes a window of 15 days of historical data. The state tensor has shape $(n \times w \times f) = (29 \times 15 \times 12)$, which includes adjusted close price, volume, and technical indicators (including EMA, MACD, RSI, and Bollinger Bands).

\paragraph{Implementation.}
Training was conducted on an HPC cluster using four NVIDIA RTX 4090 GPUs with PyTorch and the TD3 algorithm for policy optimization. Each configuration was trained for 500 episodes, requiring approximately 80 GPU-hours. We note that while training benefits from parallel GPU acceleration, inference requires minimal computational resources and can be performed on a single CPU in real-time for practical deployment. Hyperparameters were optimized using Bayesian optimization via Tree-structured Parzen Estimator (TPE) \cite{bergstra2011algorithms} over 40 trials, with search spaces detailed in Table~\ref{tab:hyperparams}. For model selection, we trained on the training set, selected the best checkpoint based on validation set Sharpe ratio, and report final performance on the held-out test set.

\paragraph{Hyperparameters.}
Table~\ref{tab:hyperparams} presents the key hyperparameters from Trial 30, which is the best-performing configuration identified through Optuna optimization.

\begin{table}[t]
    \centering
    \caption{Key hyperparameters from Optuna optimization}
    \label{tab:hyperparams}
    \resizebox{\columnwidth}{!}{%
    \begin{tabular}{llcr}
        \toprule
        Category & Parameter & Search Range & Value \\
        \midrule
        \multirow{4}{*}{Training} & Batch size & $\{128, 256, 512, 1024\}$ & 1024 \\
        & Actor LR & $[10^{-6}, 10^{-3}]$ & 2.58e-4 \\
        & Critic LR & $[10^{-5}, 10^{-2}]$ & 6.21e-4 \\
        & Discount ($\gamma$) & $[0.98, 0.999]$ & 0.991 \\
        \midrule
        \multirow{3}{*}{Transformer} & $d_{\text{model}}$ & $\{32, 64, 96, 128\}$ & 128 \\
        & Num heads & $\{2, 4, 8\}$ & 2 \\
        & Num layers & $[1, 4]$ & 4 \\
        \midrule
        \multirow{2}{*}{Critic} & Num layers & $[2, 5]$ & 3 \\
        & Hidden size & $\{256, 512, 1024\}$ & 512 \\
        \midrule
        \multirow{3}{*}{BAVAR} & Max models & $[400, 650]$ & 600 \\
        & Burn-in ratio & $[0.2, 0.4]$ & 0.34 \\
        & $\tau_{\text{BAVAR}}$ & $[0.01, 0.1]$ & 0.077 \\
        \midrule
        \multirow{2}{*}{Black-Litterman} & $\tau_{\text{BL}}$ & $[0.01, 0.1]$ & 0.039 \\
        & $\sigma^2_\Omega$ & $[0.01, 0.2]$ & 0.052 \\
        \bottomrule
    \end{tabular}}
\end{table}

\subsection{Baseline Comparisons}

We compare BAVAR-BLED against four categories of baselines using standard portfolio performance metrics: the Sharpe ratio (SR) \cite{sharpe1994sharpe}, Sortino ratio (SoR) \cite{sortino1994performance}, accumulated returns (AR), maximum drawdown (MDD) \cite{magdon2004maximum}, and annualized volatility. Classic portfolio methods include Equal Weight (1/N), Mean-Variance optimization, and Minimum Variance portfolios. Algorithmic strategies comprise Momentum and Dual Moving Average approaches. We also compare against state-of-the-art methods from two domains: deep reinforcement learning approaches including PPO and A2C from Stable-Baselines3 \cite{raffin2021stable}, EIIE \cite{jiang2017deep}, and Risk-Adjusted DRL \cite{choudhary2025risk}; and transformer-based time series methods including TFT \cite{lim2021temporal}, PatchTST \cite{nie2023patchtst}, Informer \cite{zhou2021informer}, iTransformer \cite{liu2024itransformer}, TimesNet \cite{wu2023timesnet}, and TimeXer \cite{wang2024timexer}.

\subsection{Results}

\begin{figure}[t]
    \centering
    \includegraphics[width=\columnwidth]{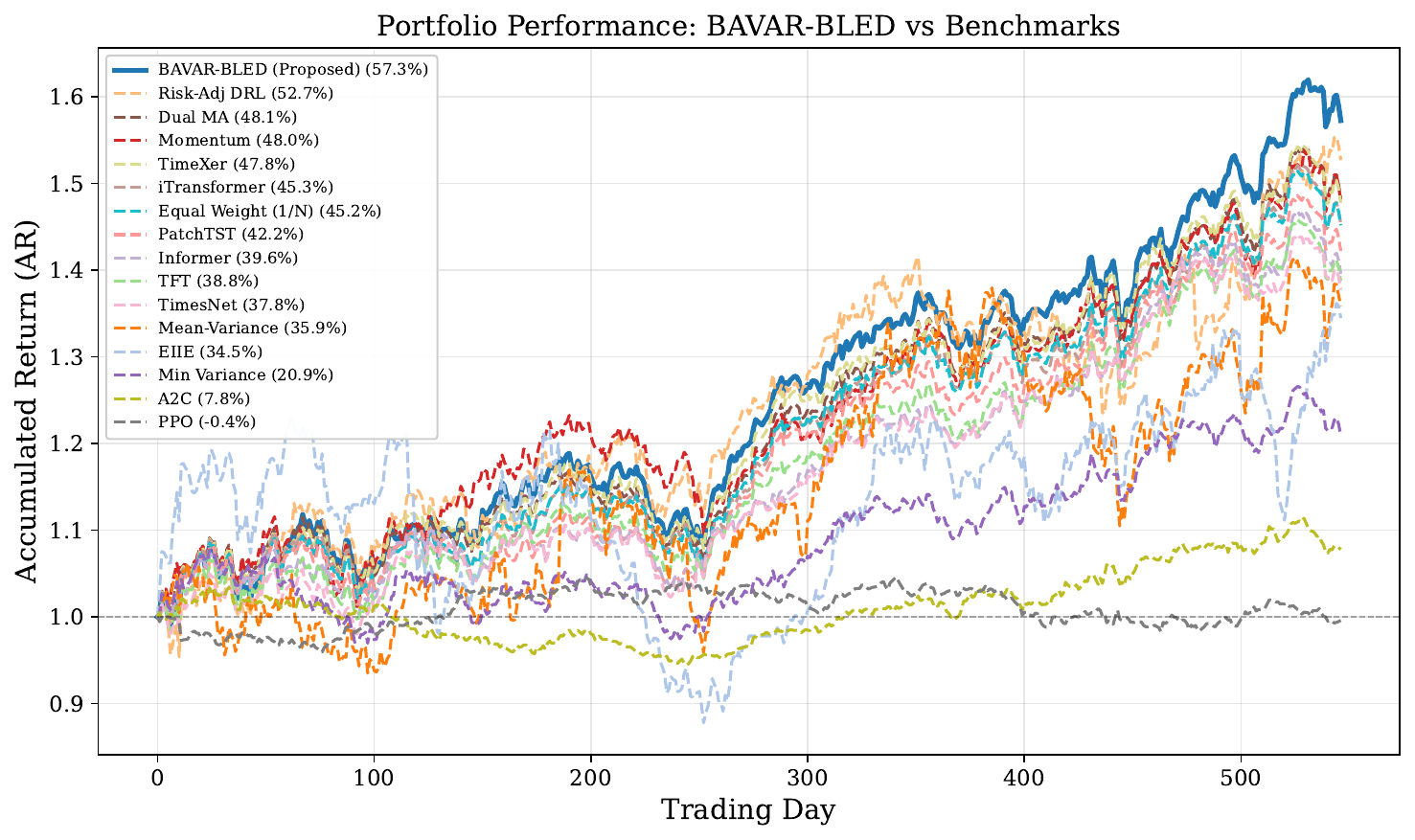}
    \caption{Cumulative returns over the test period comparing BAVAR-BLED against benchmark methods. Our method (solid blue) achieves the highest final return of 57.3\% while maintaining stable growth throughout different market conditions.}
    \label{fig:benchmark}
\end{figure}

Table~\ref{tab:results} presents the performance comparison on the test set, with cumulative returns visualized in Figure~\ref{fig:benchmark}. BAVAR-BLED achieves the highest Sharpe ratio of 1.72 among all methods while maintaining competitive volatility at 12.64\%. The proposed method outperforms the best transformer baseline (TimeXer) by 9.6\% in Sharpe ratio and 19.9\% in total returns. Compared to deep RL methods, BAVAR-BLED shows clear improvements, with Risk-Adjusted DRL achieving only 1.21 Sharpe ratio despite higher raw returns due to elevated volatility. Traditional methods such as Mean-Variance and EIIE suffer from drawdowns exceeding 19\% and 28\% respectively, while BAVAR-BLED maintains a maximum drawdown of 8.85\%. Although PPO achieves the lowest MDD and volatility, it produces negative returns and near-zero risk-adjusted performance. We note that BAVAR-BLED does not achieve the lowest volatility partially because the trading environment requires each model to invest its full available balance at every step. Models with higher cumulative wealth therefore have higher absolute volatility, as larger capital bases amplify return fluctuations. Since BAVAR-BLED maintained the highest balance for over 95\% of trading days, its volatility metric is naturally higher, and thanks to the ensemble's ability to dynamically adjust the weight of each model based on its forecasting accuracy, the BAVAR module is able to detect regime shifts, thus reducing the losses during market downturns. This regime-adaptation occurs continuously without requiring explicit regime detection or labeling, further contributing to the robustness of our proposed approach. This can be observed in Figure~\ref{fig:benchmark} around trading days 250, 430, and 500, where BAVAR-BLED shows smaller relative declines compared to most baselines.

To validate these results, we conducted binomial tests for cumulative dominance and both paired $t$-tests and sign tests for rolling Sharpe ratio comparisons across 30, 60, and 90-day windows (Table~\ref{tab:stat_tests}). The cumulative dominance test checks whether BAVAR-BLED's AR exceeds each benchmark for more than 50\% of trading days. Results show statistical significance ($p < 0.05$) against 15 of 16 benchmarks. The exception is Risk-Adjusted DRL, where BAVAR-BLED was ahead for 49.1\% of trading days ($p = 0.68$), indicating comparable AR trajectories. However, Risk-Adjusted DRL achieves this with much higher volatility (17.46\% vs 12.64\%) and worse maximum drawdown (12.83\% vs 8.85\%), resulting in inferior risk-adjusted performance as confirmed by the rolling Sharpe comparison ($p < 0.001$). For the rolling Sharpe ratio comparisons, statistical significance is determined by either the paired $t$-test or sign test across any window. In this test, all 16 benchmarks show significance. TimeXer shows a sign test $p$-value of 0.051, approaching 0.05, while the paired $t$-test remains significant ($p < 0.001$). For Momentum, the sign test is not significant ($p = 0.15$), indicating that BAVAR-BLED does not have higher rolling Sharpe for the majority of trading windows. However, the paired $t$-test is highly significant ($p < 0.001$), meaning that when BAVAR-BLED outperforms, the magnitude of outperformance exceeds the periods when it underperforms. Both methods still trail BAVAR-BLED in final returns (48.03\% and 45.41\% vs 57.26\%) and overall Sharpe ratio (1.44 and 1.53 vs 1.72), confirming the robustness of our approach.

\begin{table}[t]
    \centering
    \caption{Performance comparison on the test set (best results in bold)}
    \label{tab:results}
    \resizebox{\columnwidth}{!}{%
    \begin{tabular}{@{}lrrrrr@{}}
        \toprule
        Method & Ret. (\%) & Sharpe & Sortino & MDD (\%) & Vol. (\%) \\
        \midrule
        \textbf{BAVAR-BLED} & \textbf{57.26} & \textbf{1.72} & \textbf{2.70} & -8.85 & 12.64 \\
        \midrule
        Dual MA & 48.14 & 1.59 & 2.53 & -9.22 & 11.86 \\
        TimeXer & 47.77 & 1.57 & 2.48 & -8.49 & 11.90 \\
        Informer & 39.59 & 1.52 & 2.50 & -6.89 & 10.49 \\
        Equal Weight & 45.16 & 1.51 & 2.39 & -8.45 & 11.84 \\
        PatchTST & 42.21 & 1.50 & 2.34 & -7.56 & 11.28 \\
        iTransformer & 45.33 & 1.45 & 2.30 & -8.88 & 12.44 \\
        Momentum & 48.03 & 1.44 & 2.19 & -11.08 & 13.21 \\
        TimesNet & 37.76 & 1.44 & 2.37 & -7.54 & 10.70 \\
        TFT & 38.75 & 1.42 & 2.30 & -7.52 & 11.05 \\
        Risk-Adj DRL & 52.70 & 1.21 & 1.70 & -12.83 & 17.46 \\
        Min Variance & 20.92 & 0.92 & 1.39 & -10.33 & 10.06 \\
        EIIE & 34.49 & 0.73 & 1.04 & -28.61 & 22.24 \\
        Mean-Variance & 35.86 & 0.72 & 1.11 & -19.91 & 23.46 \\
        A2C & 7.78 & 0.65 & 1.07 & -8.66 & 5.58 \\
        PPO & -0.39 & -0.01 & -0.02 & \textbf{-6.03} & \textbf{4.91} \\
        \bottomrule
    \end{tabular}%
    }
\end{table}

\begin{table}[t]
    \centering
    \caption{Statistical significance tests. Cumulative dominance uses a binomial test for whether BAVAR-BLED is ahead $>$50\% of days. Rolling Sharpe uses paired $t$-tests and sign tests across 30, 60, and 90-day windows.}
    \label{tab:stat_tests}
    \resizebox{\columnwidth}{!}{%
    \begin{tabular}{@{}lccccc@{}}
        \toprule
        Benchmark & Days Ahead & \% Ahead & Dom. $p$ & R.Sharpe $p$ & Sig. \\
        \midrule
        Mean-Variance & 520/546 & 95.2\% & $<$0.001 & $<$0.001 & YES \\
        Min-Variance & 512/546 & 93.8\% & $<$0.001 & $<$0.001 & YES \\
        Equal Weight & 505/546 & 92.5\% & $<$0.001 & $<$0.001 & YES \\
        Momentum & 332/546 & 60.8\% & $<$0.001 & $<$0.001$^\dagger$ & YES \\
        Dual MA & 470/546 & 86.1\% & $<$0.001 & $<$0.001 & YES \\
        EIIE & 412/546 & 75.5\% & $<$0.001 & $<$0.001 & YES \\
        Risk-Adj DRL & 268/546 & 49.1\% & 0.681 & $<$0.001 & YES$^*$ \\
        PPO & 544/546 & 99.6\% & $<$0.001 & $<$0.001 & YES \\
        A2C & 542/546 & 99.3\% & $<$0.001 & $<$0.001 & YES \\
        TFT & 544/546 & 99.6\% & $<$0.001 & $<$0.001 & YES \\
        PatchTST & 545/546 & 99.8\% & $<$0.001 & 0.006 & YES \\
        Informer & 526/546 & 96.3\% & $<$0.001 & 0.047 & YES \\
        iTransformer & 543/546 & 99.5\% & $<$0.001 & 0.004 & YES \\
        TimesNet & 546/546 & 100.0\% & $<$0.001 & 0.023 & YES \\
        TimeXer & 526/546 & 96.3\% & $<$0.001 & 0.051$^\ddagger$ & YES \\
        \bottomrule
        \multicolumn{6}{l}{\scriptsize $^*$Cumulative dominance not significant; rolling Sharpe significant.} \\
        \multicolumn{6}{l}{\scriptsize $^\dagger$Sign test not significant ($p=0.15$); $t$-test highly significant.} \\
        \multicolumn{6}{l}{\scriptsize $^\ddagger$Sign test approaching significance ($p=0.05$); $t$-test significant.} \\
    \end{tabular}%
    }
\end{table}

\subsection{Ablation Study}

\begin{figure}[t]
    \centering
    \includegraphics[width=\columnwidth]{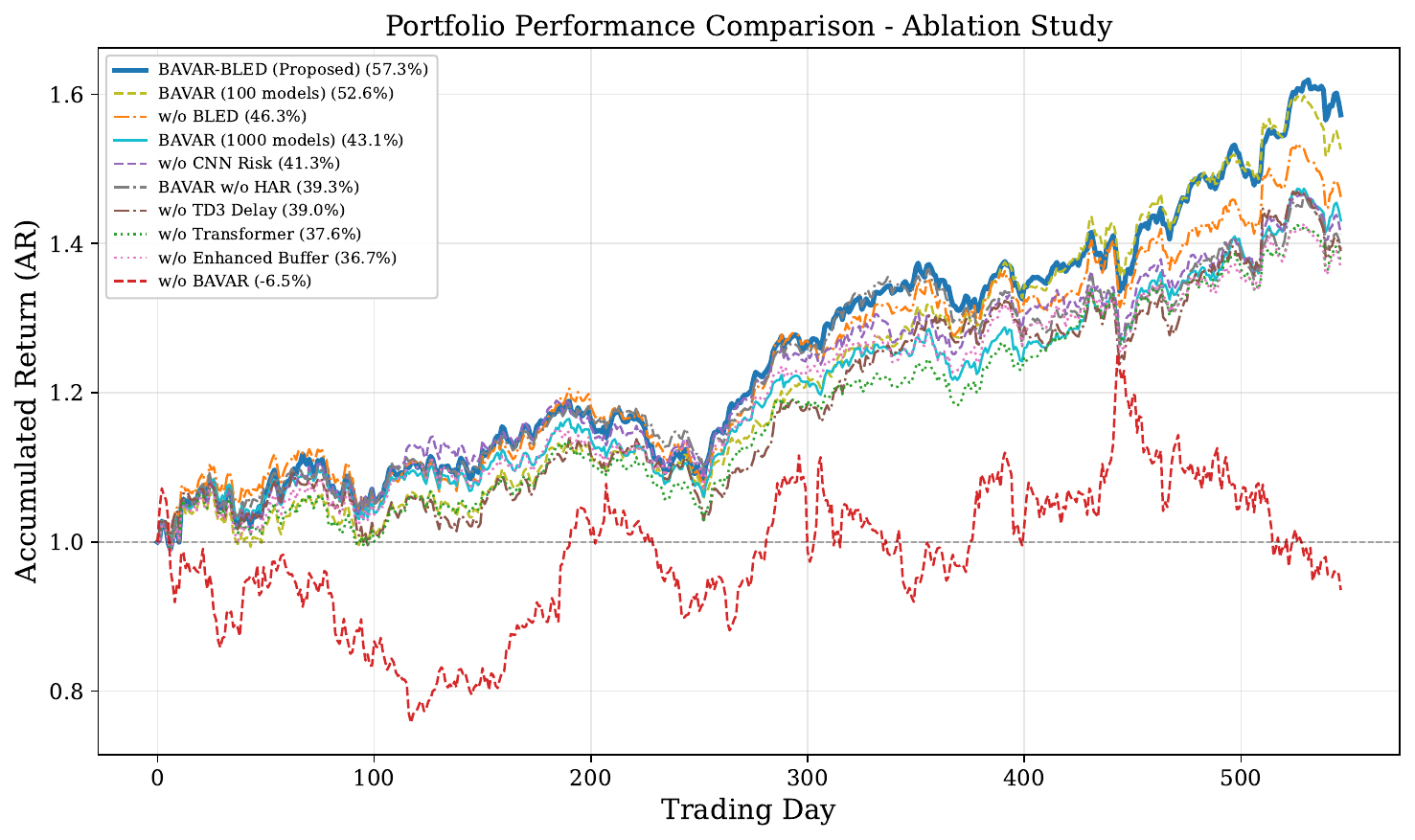}
    \caption{Cumulative returns comparison for ablation study configurations over the test period.}
    \label{fig:ablation}
\end{figure}

To understand the contribution of each component, we conducted ablation experiments by systematically removing or modifying individual framework elements. Figure~\ref{fig:ablation} shows the cumulative returns over the test period for each configuration, and Table~\ref{tab:ablation} summarizes the quantitative results.

\begin{table}[t]
    \centering
    \caption{Ablation study results showing the impact of removing individual components}
    \label{tab:ablation}
    \resizebox{\columnwidth}{!}{%
    \begin{tabular}{@{}lrrrr@{}}
        \toprule
        Configuration & Ret. (\%) & Sharpe & Sortino & MDD (\%) \\
        \midrule
        \textbf{Full BAVAR-BLED} & \textbf{57.26} & \textbf{1.72} & \textbf{2.70} & -8.85 \\
        \midrule
        BAVAR 100 models & 52.62 & 1.68 & 2.67 & -7.08 \\
        w/o BLED & 46.26 & 1.41 & 2.20 & -11.40 \\
        BAVAR 1000 models & 43.11 & 1.48 & 2.27 & -8.96 \\
        w/o CNN risk est. & 41.28 & 1.35 & 2.08 & -8.78 \\
        w/o HAR features & 39.28 & 1.39 & 2.12 & -7.57 \\
        w/o TD3 delay & 38.99 & 1.22 & 1.93 & -9.69 \\
        w/o Transformer views & 37.59 & 1.33 & 2.03 & -9.63 \\
        w/o Enhanced buffer & 36.67 & 1.25 & 1.91 & \textbf{-6.86} \\
        w/o BAVAR & -6.45 & 0.05 & 0.07 & -29.33 \\
        \bottomrule
    \end{tabular}%
    }
\end{table}

The ablation results reveal several important findings. First, removing BAVAR causes the model to fail, with returns dropping to -6.45\% and the Sharpe ratio falling to 0.05, demonstrating that the adaptive prior estimation is essential for the framework. Second, removing BLED also leads to a notable performance drop (Sharpe from 1.72 to 1.41). As stated previously, financial returns exhibit fatter tails \cite{ellip3,ellip2,ellip1}, which makes modeling with Student's $t$-distributions avoid the underestimation of extreme event probabilities, thus leading to more appropriate position sizing during volatile periods, as shown in Figure~\ref{fig:benchmark}. Third, as it can be seen from Table~\ref{tab:ablation}, both higher (1000) and lower (100) number of BAVAR models negatively impacted the performance of the framework, which can be explained by the bias-variance trade-off in Bayesian model averaging, where few models provide insufficient diversity \cite{dormann2018model} to capture market movements, while more models can introduce noise from poorly performing models and dilute the signal from more well-calibrated models \cite{jay2021bayesian}, thus further confirming the findings of the hyperparameter tuning and validating 600 models as being a sweet spot for the BAVAR-BLED framework. Fourth, transformer-based view generation and CNN risk estimation both contribute meaningfully, with their removal causing 23\% and 22\% reductions in the Sharpe ratio respectively. Finally, HAR features improve performance by capturing multi-scale temporal dynamics, and TD3 architectural choices (delayed updates, enhanced replay buffer) provide additional stability gains. Similarly, the configuration without enhanced buffer achieves the lowest MDD of 6.86\%, but this comes with a 27\% decrease in Sharpe ratio, indicating that the full framework provides a better trade-off between risk control and returns.

\section{Conclusion}

This paper presented BAVAR-BLED, a framework for portfolio optimization that addresses two fundamental limitations in existing DRL approaches: the assumption of normally distributed returns by utilizing the Black-Litterman model under Elliptical Distributions and the uniform treatment of historical data by introducing Bayesian-Averaging Vector Autoregressive methods. In this way, our approach provides adaptive, regime-aware prior estimates that respond to changing market conditions while modeling fat-tailed return distributions more realistically. Furthermore, an experimental evaluation of 29 DJIA stocks demonstrates that BAVAR-BLED achieves a Sharpe ratio of 1.72 and total returns of 57.26\%, outperforming both traditional portfolio methods and state-of-the-art transformer-based approaches, thereby establishing a new performance benchmark in this domain. These results were further validated through statistical tests for AR and rolling SR comparisons, confirming the robustness of the framework and its adaptability to changing market conditions. The ablation study confirms that each component contributes meaningfully to performance, with BAVAR being essential to the framework, as its removal reduces the Sharpe ratio from 1.72 to 0.05.

The main limitations of our approach include computational complexity from maintaining a large VAR ensemble and restricted HAR feature horizons. While the original HAR methodology \cite{corsi2009har} incorporated yearly and 10-year horizons, our implementation uses only daily, weekly, and monthly (1, 5, 22 trading days) features due to computational constraints. This may limit the model's ability to capture longer-term market cycles.

Future work will explore several directions. First, the framework naturally supports continuous learning during deployment, as BAVAR updates through closed-form Bayesian equations rather than gradient descent, allowing for adaptation to new market observations without retraining the computationally expensive DRL component. Second, extending HAR features to longer horizons like quarterly, yearly and 10-year could improve the detection of long-term market trends. Lastly, incorporating transfer learning and optimizing ensemble management could reduce the burn-in period for new market environments.

\bibliographystyle{named}
\bibliography{references}

\end{document}